\documentclass[11pt]{article}

\usepackage{amsmath,amsfonts,bm}

\def\eqref#1{equation~\ref{#1}}

\def\1{\bm{1}}

\DeclareMathAlphabet{\mathsfit}{\encodingdefault}{\sfdefault}{m}{sl}
\SetMathAlphabet{\mathsfit}{bold}{\encodingdefault}{\sfdefault}{bx}{n}

\DeclareMathOperator{\sign}{sign}

\usepackage{booktabs}
\usepackage{luca}
\usepackage{graphicx}
\usepackage{amsfonts,amsmath,amssymb}
\usepackage{url}
\usepackage{hyperref}
\usepackage{comment} 
\usepackage{float} 
\usepackage{tablefootnote} 
\usepackage{xcolor}
\usepackage{geometry} 
\usepackage{rbf}
\usepackage{subfigure}
\newlength\figsize
\begin{document}
	\title{A New Family of Neural Networks Provably Resistant
		to Adversarial Attacks}

	\author{Rakshit Agrawal \quad Luca de Alfaro \quad David Helmbold\thanks{The authors are listed in alphabetical order.} 
	\\[1ex]
		\normalsize Computer Science and Engineering Department\\
		\normalsize University of California, Santa Cruz\\
		\normalsize ragrawa1@ucsc.edu, luca@ucsc.edu, dph@ucsc.edu}
	\date{January 31, 2019 \\
		\normalsize Technical Report UCSC-SOE-19-01\\
		\normalsize School of Engineering, University of California, Santa Cruz}
	\maketitle
	
	\def\citet#1{\cite{#1}}
	
	\begin{abstract}
	Adversarial attacks add perturbations to the input features with the intent of changing the classification produced by a machine learning system. 
Small perturbations can yield adversarial examples which are misclassified despite being virtually indistinguishable from the unperturbed input.  
Classifiers trained with standard neural network techniques are highly susceptible to adversarial examples, allowing an adversary to create misclassifications of their choice. 


We introduce a new type of network unit, called MWD (max of weighed distance) units that 
have a built-in resistant to adversarial attacks.
These units are highly non-linear, and we develop the techniques needed to effectively train them. 
We show that simple interval techniques for propagating perturbation effects through the network enables the efficient computation of robustness (i.e., accuracy guarantees) 
for MWD networks under any perturbations, including adversarial attacks. 

MWD networks are significantly more robust to input perturbations than ReLU networks.  
On permutation invariant MNIST, when test examples can be perturbed by 20\% of the input range, MWD networks provably retain accuracy above 83\%, while the accuracy of ReLU networks drops below 5\%.
The provable accuracy of MWD networks is superior even to the observed accuracy of ReLU networks trained with the help of adversarial examples.
In the absence of adversarial attacks, MWD networks match the performance of sigmoid networks, and have accuracy only slightly below that of ReLU networks.

	\end{abstract}
	
	\section{Introduction}
\label{s-intro}

Machine learning via deep neural networks has been remarkably successful in a wide range of applications, including speech recognition, image classification, and language processing. 
While very successful, deep neural networks are susceptible to  adversarial examples: small, carefully crafted, perturbations of inputs can change their predicted classification
 \cite{SzegedyIntriguingpropertiesneural2013}. 
These perturbations can often be so small as to make human detection difficult or impossible; this has been shown both in the case of images \cite{SzegedyIntriguingpropertiesneural2013,NguyenDeepneuralnetworks2015a} and sounds \cite{KurakinAdversarialexamplesphysical2016,CarliniAudioadversarialexamples2018}. 
Further, the adversarial examples are in some sense transferable from one neural network to another (\cite{GoodfellowExplainingharnessingadversarial2014,NguyenDeepneuralnetworks2015a,Papernotlimitationsdeeplearning2016,Tramerspacetransferableadversarial2017}), so they can be crafted even without precise knowledge of the targeted network's parameters.
At a fundamental level, it is hard to be confident about the behavior of a deep neural network when most correctly classified inputs are in close proximity to very similar, yet differently classified inputs. 

The dominant approach for increasing a neural network's resistance to adversarial attacks is to augment the training data with adversarial examples \cite{GoodfellowExplainingharnessingadversarial2014,MadryDeepLearningModels2017}.
If the added adversarial examples are generated by efficient heuristics such as the {\em fast gradient sign method,} the networks learn to associate the specific adversarial examples with the corresponding unperturbed input in a phenomenon known as {\em label leaking\/} \cite{KurakinAdversarialmachinelearning2016,MadryDeepLearningModels2017,TramerEnsembleadversarialtraining2017}. 
This does not result in increased resistance to more general adversarial attacks \cite{MadryDeepLearningModels2017,Carlinievaluatingrobustnessneural2017}. 
Networks with greater resistance to adversarial attacks can be obtained if the adversarial examples used in training are generated via more general optimization techniques, as in \cite{MadryDeepLearningModels2017}.
This comes at the cost of a more complex and computationally expensive training regime, as well as an increase in network's capacity.

Here we propose a different approach, the use of neural network units that are inherently resistant to adversarial attacks, even when trained using only unperturbed input.
\citet{GoodfellowExplainingharnessingadversarial2014} connect the presence of adversarial examples to the (local) linearity of neural networks. 
Consider the linear form $\sum_{i=1}^n x_i w_i$ and perturbing each $x_i$ by $\epsilon$, taking $x_i + \epsilon$ if $w_i > 0$, and $x_i - \epsilon$ if $w_i < 0$.
The output is then perturbed by $\epsilon \sum_{i=1}^n |w_i|$, or $\epsilon n \bar{w}$ where $\bar{w}$ the average magnitude of the $w_i$'s.
When the number of inputs $n$ is large, as is typical of deep neural networks, a small input perturbation can cause a large output change, and this change can snowball through the layers.
Of course, deep neural networks are not globally linear, but the insight of \citet{GoodfellowExplainingharnessingadversarial2014} is that they may be sufficiently locally linear to promote the success of adversarial attacks. 
Following this insight, we develop networks composed of units that are highly non-linear.

After much experimentation, we found a promising node type that we call 
\emph{Max-of-Weighted-Distance} (MWD) units. 
Like Gaussian radial basis functions \cite{BroomheadRadialbasisfunctions1988,ChenOrthogonalleastsquares1991,OrrIntroductionradialbasis1996},
MWD units activate based on the distance from their input vectors to a learned center.
However, MWD units also learn a non-negative weight for each input component 
and measure distance from the center with the  weighted infinity-norm rather than the Euclidean norm.
The component weighing can give a high sensitivity to some components while ignoring others, allowing a single MWD unit to cover a more flexible region of the input space.
The use of the infinity norm reduces sensitivity to adversarial perturbations because any change in the output is due to the perturbation of one input component, rather than the sum
of the perturbations to all of the input components.
The output of a MWD unit $\neuron$ with parameters $\vecu$ and $\vecw$ on input $\vecx$ is
\begin{equation} \label{eq-rbfinf}
  \neuron(\vecu, \vecw)(\vecx) = \exp\Bigl(- \max_{1 \leq i \leq n} \bigl(u_i (x_i - w_i)\bigr)^2 \Bigr) \eqpun . 
\end{equation}

Using highly nonlinear models is hardly a new idea, but the challenge has been that such models are typically difficult to train. 
Indeed, we found that networks with MWD units cannot be satisfactorily trained using gradient descent. 
To get around this, we show that the networks can be trained efficiently, and to high accuracy, using {\em pseudogradient descent} where
the {\em pseudogradient\/} is a proxy for the gradient that facilitates training. 
The maximum operator in~(\ref{eq-rbfinf}) has non-zero derivative only for the maximizing input; our pseudogradient propagates a derivative signal back to all of the inputs.
Also, the exponential function in~(\ref{eq-rbfinf}) is very flat far away from the center, so our pseudogradient artificially widens the region of meaningful gradients. 
Tampering with the gradient may seem unorthodox, but methods such as AdaDelta (\cite{ZeilerADADELTAadaptivelearning2012}), 
and even gradient descent with momentum, cause training to take a trajectory that does not follow pure gradient descent. 
We simply go one step further, devising a scheme that operates at the granularity of the individual unit.

In order to prove accuracy bounds for MWD networks, we rely on the propagation of perturbation intervals thrugh the network. 
These are conservative estimates to the set of values that can be produced at a network unit, when the network inputs are subject to perturbation bounded by a specified amount. 
These technique can be used to prove assertions such as ``For any input perturbations of size at most $x$ in infinity norm, the network yields accuracy of at least y\% on the testing set''. 
The guarantees {\em are\/} dependent on the testing set, and in general, on the probability distribution over the inputs, but this is intrinsic to any guarantee or performance measure in machine learning. 
ReLU networks, even when trained with the help of adversarial examples, can offer only very low accuracy guarantees. 
In contrast, we show that networks of MWD units, even when trained normally, offer performance lower-bounds under attack that are superior even to the (upper bounds) provided by 
ReLU networks trained with adversarial examples.

To conduct our experiments, we have implemented MWD networks on top of the PyTorch framework \cite{PaszkeAutomaticdifferentiationpytorch2017}.
The code for the MWD networks is available at \url{https://github.com/rakshit-agrawal/mwd_nets}.
We consider {\em permutation invariant MNIST,} which is a version of MNIST in which the $28 \times 28$ pixel images are flattened into a one-dimensional vector of $784$ values and fed as a feature vector to neural networks \cite{GoodfellowExplainingharnessingadversarial2014}.
On this test set, we show that for nets of 512,512,512,10 units, MWD networks match the classification accuracy of sigmoid networks ($(96.96 \pm 0.14)\%$ for MWD vs.\ $(96.88 \pm 0.15)\%$ for sigmoid), and are close to the performance of network with ReLU units ($(98.62 \pm 0.08)\%$). 
When trained over standard training sets, for input perturbations of 20\% of the input range, MWD networks guarantee an accuracy above 80\% for any adversarial attack, while there are simple attacks that reduce the accuracy  of ReLU and Sigmoid networks to below 10\% (random guessing). 
Even when ReLU networks are trained with the benefit of adversarial attacks, for the most relevant range of input perturbations (from 5\% to 25\% of the input range), the accuracy lower bound guarantees offered by (normally trained) MWD networks exceed the upper bounds of ReLU networks due to known attacks. 

Our results can be summarized as follows:
\begin{itemize}
\item We define a class of networks, MWD networks, inherently resistant to adversarial attacks, and we develop a pseudogradient-based way for effectively training them.
\item On MNIST, we show that in absence of adversarial attacks, MWD networks match the accuracy of sigmoid networks, and have only slightly lower accuracy than ReLU networks.
\item Again on MNIST, in presence of adversarial attacks, we show that the accuracy lower-bound guarantees of MWD networks far exceed the accuracy upper bounds of ReLU and Sigmoid networks. 
We show that the accuracy lower bounds of MWD networks are above the ReLU upper bounds even when the latter are trained with the help of adversarial examples.
\end{itemize}
Much work remains to be done, including experimenting with MWD units in convolutional networks. 
However, these initial results offer a practical method for training networks that are provably --- and significantly -- resistant to adversarial attacks.

	\section{Related Work}

The vulnerability of neural networks to adversarial examples was first reported by \citet{SzegedyIntriguingpropertiesneural2013}, and they showed
how to generate them via a simple optimization.
\citet{GoodfellowExplainingharnessingadversarial2014} established a connection between linearity and adversarial attacks. 
A fully linear form $\sum_{i=1}^n x_i w_i$ can be perturbed to $x_i + \epsilon \, \sign(w_i)$, creating an output change of magnitude $\epsilon \cdot \sum_{i=1}^n |w_i|$. 
In analogy, \citet{GoodfellowExplainingharnessingadversarial2014} introduced the {\em fast gradient sign method} (FGSM) method of creating adversarial perturbations, 
by taking $x_i + \epsilon \cdot \sign(\grad_i \call)$, where $\grad_i \call$ is the loss gradient with respect to input $i$. 
They also showed how adversarial examples are often transferable across networks, and asked if non-linear structures, perhaps like those of RBFs, would be more robust to adversarial attacks. 
This paper pursues this approach and provides positive answers to the conjectures and suggestions by \citet{GoodfellowExplainingharnessingadversarial2014}. 

It was recently discovered that training on adversarial examples generated via FGSM does not confer strong resistance to attacks, as the network learns to associate the specific examples generated by FGSM with the original training examples in a phenomenon known as {\em label leaking\/} \cite{KurakinAdversarialmachinelearning2016,MadryDeepLearningModels2017,TramerEnsembleadversarialtraining2017}.
The FGSM method for generating adversarial examples was extended to an iterative method, I-FGSM, in \cite{KurakinAdversarialexamplesphysical2016}.
In \cite{TramerEnsembleadversarialtraining2017}, it is shown that using small random perturbations before applying FSGM enhances the robustness of the resulting network. 
The network trained in \cite{TramerEnsembleadversarialtraining2017} using I-FSGM and ensemble method won the first round of the NIPS 2017 competition on defenses with respect to adversarial attacks. 

Carlini and Wagner show that training regimes based on generating adversarial examples via simple heuristics, or combinations of these, in general fail to convey true resistance to attacks \cite{CarliniAdversarialexamplesare2017,Carlinievaluatingrobustnessneural2017}.
They further advocate measuring the resistance to attacks with respect to adversarial examples created by more general optimization processes. 
In particular, FGSM and I-FGSM rely on the local gradient, and training techniques that break the association between the local gradient and the location of adversarial examples makes networks harder to attack via FGSM and I-FGSM, without making the networks harder to attack via general optimization techniques. 
We follow this suggestion by using a general optimization method, projected gradient descent (PGD), to generate adversarial attacks and evaluate network robustness. 
\citet{CarliniDefensivedistillationnot2016,Carlinievaluatingrobustnessneural2017} also show that the technique of {\em defensive distillation,} which consists in appropriately training a neural network on the output of another \cite{PapernotDistillationdefenseadversarial2016}, protects the networks from FGSM and I-FGSM attacks, but does not improve network resistance in the face of general adversarial attacks. 

\citet{MadryDeepLearningModels2017} show that it is possible to obtain networks that are genuinely more resistant to adversarial examples by training on adversarial examples generated via PGD.
The price to pay is a more computationally intensive training, and an increase in the network capacity required. 
We provide an alternative way of achieving such resistance that does not rely on a new training regime. 

A technique based on differential analysis for deriving lower bounds to resistance to adversarial attacks is presented in \cite{NIPS2017_6682}. 
The technique requires the feed-forward functions to be (locally) differentiable, and therefore does not provide bound guarantees for networks including non-differentiable units, such as ReLU and MWD networks. 
\cite{EhlersFormalverificationpiecewise2017} use an interval propagation technique for bounding the range of ReLU node activations 
before applying ILP and SAT-solver techniques to verify network robustness.  
Our results show that for MWD networks, the simple technique of forward propagating perturbation intervals gives meaningful robustness guarantees. 

	\section{MWD Networks}
\label{sec-rbfi}


\citet{GoodfellowExplainingharnessingadversarial2014} link adversarial attacks to the linearity of the models. 
In a linear form $g(\vecx) = \sum_i x_i w_i$, if we perturb $x_i$ by adding $\epsilon$ when $w_i > 0$, and subtracting it when $w_i < 0$, the perturbations on the various inputs add up, so that small input perturbations can yield large output changes. 
To achieve resistance to adversarial attacks, we seek units where the contributions of the inputs are not added up. 
The linear form represents the norm-2 distance of the input vector $x$ to a hyperplane perpendicular to vector $\vecw$, scaled by $|\vecw|$ and its orientation.  
Our units will be based instead on infinity-norm distances. 

We define our units as variants of the classical Gaussian {\em radial basis functions\/} \cite{BroomheadRadialbasisfunctions1988,OrrIntroductionradialbasis1996}. 
We call our variant \emph{Max-of-Weighted-Distance\/} (\rbfinf), to emphasize the fact that they are built using infinity norm.
An \rbfinf\ unit $\neuron(\vecu, \vecw)$ for an input in $\reals^n$ is parameterized by two vectors of weights $\vecu = \tuple{u_1, \ldots, u_n}$ and $\vecw = \tuple{w_1, \ldots, w_n}$
Given an input $\vecx \in \reals^n$, the unit produces output 
\begin{equation} \label{eq-rbf}
  \neuron(\vecu, \vecw)(\vecx) = \exp\left(- \norm{\infty}{\vecu \hadamard (\vecx - \vecw)}^2\right) \eqpun ,
\end{equation}
where $\hadamard$ is the Hadamard, or element-wise, product.
In (\ref{eq-rbf}), the vector $\vecw$ is a point from which the distance to $\vecx$ is measured in infinity norm, and the vector $\vecu$ provides independent scaling factors for each coordinate.
Without loss of expressiveness, we require the scaling factors to be non-negative, that is, $u_i \geq 0$ for all $1 \leq i \leq n$. 
The scaling factors provide the flexibility of disregarding some inputs $x_i$, by having $u_i \approx 0$, while emphasizing the influence of other inputs. 
Expanding (\ref{eq-rbf}) gives: $\neuron(\vecu, \vecw)(\vecx) = \exp\Bigl(- \max_{1 \leq i \leq n} \bigl(u_i (x_i - w_i)\bigr)^2 \Bigr)$ as in equation~(\ref{eq-rbfinf}).
%
%

The output of  \rbfinf\ units are in $(0,1]$ and are close to $1$ only when $\vecx$ is close to $\vecw$ in the coordinates that have significant scaling factors. 
Thus, the unit functions somewhat like an \textsc{and} gate, outputting $1$ only when the relevant inputs take on a particular set of values.
We also consider the negated \rbfinf\ unit which functions somewhat like a $\textsc{nand}$ gate:
$\neuron^{\mbox{\sc \tiny neg}}(\vecu, \vecw) = 1 - \neuron(\vecu, \vecw)$.
%
We construct neural networks out of \rbfinf\ units using layers consisting of $\neuron$  units, layers consisting of $\neuron^{\mbox{\sc \tiny neg}}$ units, and mixed layers where the unit type is chosen at random at network initialization.

\subsection{Training \rbfinf\ Networks via Pseudogradients}

The non-linearities in (\ref{eq-rbfinf}) make neural networks containing \rbfinf\ units difficult to train using standard gradient descent.
The problems are associated with the $\max$-operator making the gradients sparse and the fast decay of Gaussian functions.
Far from its peak for $\vecx = \vecw$, a function of the form (\ref{eq-rbfinf}) is rather flat, and its derivative may not be large enough top cause the vector of weights $\vecw$ to move towards useful places in the input space during training. 
To obtain networks that are easy to train, we replace the derivatives for $\exp$ and $\max$ with alternate functions, which we call {\em pseudoderivatives.}
These {\em pseudoderivatives\/} are then used in the chain-rule computation of the loss gradient in lieu of the true derivatives, yielding a {\em pseudogradient.}

\paragraph{Exponential function.}
In computing the partial derivatives of (\ref{eq-rbf}) via the chain rule, the first step consists in computing
$
\frac{d}{dz} e^{-z} = -e^{-z}
$ .
The problem is that $-e^{-z}$ is very close to 0 when $z$ is large, and $z$ in (\ref{eq-rbfinf}) is $\norm{\infty}{\vecu \hadamard (\vecx - \vecw)}^2$, which can be large. 
Hence, in the chain-rule computation of the gradient, we replace $-e^{-z}$ with the ``pseudoderivative'' 
$
- 1 /\sqrt{1 + z}
$, 
which decays much more slowly.
%


\paragraph{Max.}
The gradient of $y = \max_{1 \leq i \leq n} z_i$, of course, is given by $\frac{\partial y}{\partial z_i} = 1$ if $z_i = y$, and $\frac{\partial y}{\partial z_i} = 0$ if $z_i < y$. 
The problem is that this transmits feedback only to the largest input(s). 
This slows down training and can create instabilities. 
We use as pseudoderivative
$ 
  e^{z_i - y}
$, 
so that some gradient feedback is transmitted to the other inputs $z_i$ based on their closeness to $y$. 

\medskip

One may be concerned that by using the loss pseudogradient as the basis of optimization, rather than the true loss gradient, we may converge to solutions where the pseudogradient is null, and yet, we are not at a minimum of the loss function. 
This can indeed happen. 
We experimented with switching to training with true gradients after the accuracy reached via pseudogradients plateaued; this increased the  accuracy on the training set, but improved only slightly the accuracy on the testing set. 
We also experimented with pseudogradients parameterized by a parameter $\rho \in [0, 1]$, such that when $\rho=0$ the pseudogradients coincide with the true gradients.
During training the parameter $\rho$ starts at 1 and then, as training proceeds, gradually decays to~0, thus allowing a smooth transition from a pseudogradient training regime to a standard gradient one.  
At least for MNIST, these more sophisticated schemes failed to significantly improve on the simple use of the pseudogradients above.

	\newcommand{\bx}{\vecx}
\newcommand{\bu}{\vecu}
\newcommand{\bw}{\vecw}
\newcommand{\beps}{\mathbf{\epsilon}}
\newcommand{\perb}[1]{{I}_{#1}^\epsilon}
\newcommand{\perbj}{\perb{j}}

\section{Adversarial Examples}

%

\subsection{Generating Adversarial Examples}

We describe the known methods for generating candidate adversarial examples that we will use in the experiments.
Consider a network trained with cost function $J(\theta, \vecx, \vecy)$, where $\theta$ are the network parameters, $\vecx$ is the input, and $\vecy$ is the output. 
Let $\nabla_{\vecx} J(\theta, \vecx', \vecy)$ be the gradient of $J$ wrt its second argument (the input) computed at $\theta, \vecx',  \vecy$.
For a perturbation amount $\epsilon>0$ and an input $\vecx$ belonging to the test set, we produce candidate adversarial examples $\advvecx$ with $\norm{\infty}{\vecx - \advvecx} \leq \epsilon$ using the following techniques. 

\paragraph{Fast Gradient Sign Method (FGSM)} \cite{GoodfellowExplainingharnessingadversarial2014}. 
In the FGSM technique, the candidate adversarial example is generated via:
\begin{equation} \label{eq-FGSM}
  \advvecx = \clamp{0}{1}{\vecx + \epsilon \, \sign(\nabla_{\vecx} J(\theta, \vecx, \vecy))} \eqpun , 
\end{equation}
where $\clamp{a}{b}{\vecx}$ is the result of clamping each component of $\vecx$ to the range $[a, b]$; the clamping is necessary to generate a valid MNIST image.

\paragraph{Iterated Fast Gradient Sign Method (I-FGSM)} \cite{KurakinAdversarialexamplesphysical2016}.
The I-FGSM attack computes a sequence $\advvecx_0, \advvecx_1, \ldots, \advvecx_M$, where $\advvecx_0 = \vecx$, and, for $0 \leq i < M$: 
\begin{equation} \label{eq-I-FGSM}
  \advvecx_{i+1} = \clamp{0}{1}{\advvecx_i + \frac{\epsilon}{M} \, \sign(\nabla_{\vecx} J(\theta, \advvecx_i, \vecy))} \eqpun .
\end{equation}
We take $\advvecx = \advvecx_M$ as the candidate adversarial example. 

\paragraph{Projected Gradient Descent (PGD)} \cite{MadryDeepLearningModels2017}.
For an input $\vecx \in \reals^n$ and a given maximum perturbation size $\epsilon > 0$, we consider the set $B_\epsilon(\vecx) \inters [0,1]^n$ of valid perturbations of $\vecx$, and we perform projected gradient descent (PGD) in $B_\epsilon(\vecx) \inters [0,1]^n$ of the negative loss with which the network has been trained (or, equivalently, projected gradient ascent wrt.\ the loss). 
We perform this search with multiple restarts, each chosen uniformly at random from $B_\epsilon(\vecx) \inters [0,1]^n$.


\subsection{Lower Bounds via Interval Propagation}
\label{sec-bounds}
In order to obtain lower bounds for the accuracy of MWD networks subject to adversarial attacks, we propagate through the network {\em perturbation intervals\/} that represent an over-approximation of the range of values the output of a network node can assume, when the input is subject to perturbations. 
More sophisticated analysis methods exist, based on the propagation of invariants \cite{EhlersFormalverificationpiecewise2017,wong18a,RaghunathanCertifieddefensesadversarial2018,MirmanDifferentiableabstractinterpretation2018}, but for MWD, as we will see, strong bounds can be obtained simply via such interval propagation. 

The \emph{$\epsilon$-perturbation} of an input $\vecx$ is any $\vecx'$ where $|| \vecx - \vecx' ||_\infty \leq \epsilon$ and the input belongs to the input domain. 
Given a classifier, an \emph{$\epsilon$-adversarial example} for input $\vecx$ is an $\epsilon$-perturbation of $\vecx$ that results in a different prediction.
The \emph{true $\epsilon$-attack accuracy} of a trained network is the fraction of test examples for which the network predicts correctly and no $\epsilon$-adversarial examples exist.

If the network input $\vecx'$ is an $\epsilon$-perturbation of $\vecx$, then every node $j$ outputs values in its perturbation interval $\perbj(\vecx)$. 
If the network predicts correctly on some test example $\vecx$ and every combination of values within the output nodes' $\epsilon$-perturbation intervals lead to the same prediction, then $\vecx$ has no $\epsilon$-adversary examples so $\vecx$ can be counted towards the network's true $\epsilon$-attack accuracy.

In particular, consider a classification problem with $n$ classes, and a classifier network $f$ with $n$ outputs, $y_1, \ldots, y_n$. 
Input $\vecx$ (belonging to, say, class 1) is classified correctly by $\vecy = f(\vecx)$ if $y_1 > y_k$ for $k = 2, \ldots, n$. 
To rule out the existence of $\epsilon$-attacks for $\vecx$, it suffices to check that $l_1 > r_k$ for $k = 2, \ldots, n$, where $[l_k, r_k] = {\cali}_k^\epsilon(\vecx)$ is the $\epsilon$-perturbation interval for output $k = 1, \ldots, n$ and input vector $\vecx$.

For many node types the perturbation intervals are easy to compute. 
In particular, for an MWD node, let $[l_1, r_1], \ldots, [l_n, r_n]$ be the perturbation intervals of its $n$ inputs for a given $\vecx$, and denote by $f(\cdot)$ the node function, defined as by (\ref{eq-rbf}).
Its output perturbation interval $[l, r]$ for $\vecx$ can be computed via 
$l = \max_{1 \leq i \leq n} \min \bigl\{ f(l_i), f(r_i) \bigr\}$, and  
$r = \max_{1 \leq i \leq n} m_i$,
where
\[
  m_i = 
  \begin{cases}
    1 & \mbox{if $l_i \leq w_i \leq r_i$} \\
    \max \bigl\{ f(l_i), f(r_i) \bigr\} & \mbox{otherwise.} 
  \end{cases}
\]
Computing these input-dependent perturbation intervals takes effort comparable to (perhaps 2$\times$ or 3$\times$) that of calculating the network's output.

	\section{Experimental Setup}

We implemented \rbfinf\ networks in the PyTorch framework \cite{PaszkeAutomaticdifferentiationpytorch2017}.
To implement pseudogradients, we extend PyTorch with two new functions: a {\em LargeAttractorExp\/} function, with forward behavior $e^{-x}$ and backward gradient propagation according to $-1/\sqrt{1 + x}$, and {\em SharedFeedbackMax,} with forward behavior $y = \max_{i=1}^n x_i$ and backward gradient propagation according to $e^{x_i - y}$. 
These two functions are used in the definition of \rbfinf\ units, as per (\ref{eq-rbfinf}), with the AutoGrad mechanism of PyTorch providing  backward (pseudo)gradient propagation for the complete networks.

\paragraph{Dataset.}
We use the MNIST dataset \cite{LeCunGradientbasedlearningapplied1998} for our experiments, following the standard setup of 60,000 training examples and 10,000 testing examples. 
Each digit image was flattened to a one-dimensional feature vector of length $28 \times 28 = 784$, and fed to a fully-connected neural network; this is the so-called {\em permutation-invariant\/} MNIST. 

\paragraph{Neural networks.}
We compared the accuracy of ReLU, sigmoid, and \rbfinf\ networks. 
The output of ReLU networks \cite{NairRectifiedlinearunits2010} is fed into a softmax, and the network is trained via cross-entropy loss. 
Sigmoid and \rbfinf\ networks are trained via square-error loss, which worked better than other losses in our experiments.
We trained all networks with the AdaDelta optimizer \cite{ZeilerADADELTAadaptivelearning2012}, which gave good results for all networks considered.  

\paragraph{Attacks.}
We applied FGSM and I-FGSM attacks to the whole test set.
In I-FGSM attacks, we performed 10 iterations of (\ref{eq-I-FGSM}).  
As PGD attacks are computationally intensive, we apply them to one run only, and we compute the performance under PGD attacks for the first 2,000 inputs in the test set for ReLU and Sigmoid networks, and for 1,000 inputs for \rbfinf\ networks.
For each input $\vecx$ in the test set, we perform 100 PGD searches, or restarts.
Each search starts from a random point in $\ball_\epsilon(\vecx)$ and then does 100 steps of projected gradient descent using the AdaDelta algorithm to tune step size; if at any step a misclassified example is generated, the attack is considered successful. 

For \rbfinf\ networks, we can perform FGSM, I-FGSM, and PGD attacks based either on true gradient, or on the pseudogradients. 
We denote the pseudogradient-based attacks with [psd] in the figures and tables.
Such attacks in general are more powerful than attacks based on the regular gradient, for the same reasons why the pseudogradient is more effective in training.

	\section{Results}
\label{sec-results-adv}

Unless otherwise noted, we report results on networks with layers of 512, 512, 512, and 10 units. 
For \rbfinf\ networks, the layers consist of $\neuron$, $\neuron^{\mbox{\sc \tiny neg}}$, $\neuron$, $\neuron^{\mbox{\sc \tiny neg}} $ units respectively, and we use a bound of $[0.01, 3]$ for the components of the $u$-vectors, and of $[0, 1]$ for the $w$-vectors, the latter corresponding to the value range of MNIST pixels. 

\begin{table*}[t]
\centering
\begin{tabular}{r||r|r|r|r|r}
Network
 & Accuracy 
 & FGSM, $\epsilon {=} 0.3$ 
 & I-FGSM, $\epsilon {=} 0.3$
 & PGD, $\epsilon {=} 0.3$ \\ \hline
ReLU 
 & $\mathbf{98.62 \pm 0.08}$
 & $1.98 \pm 0.42$
 & $0.06 \pm 0.06$
 & $81.20$ \\
Sigmoid
 & $96.88 \pm 0.15$
 & $0.71 \pm 0.43$
 & $0.11 \pm 0.11$
 & $40.75$ \\
\rbfinf
 & $96.96 \pm 0.14$ 
 & $\mathbf{94.90 \pm 0.35}$ 
 & $\mathbf{93.27 \pm 0.48}$ 
 & $\mathbf{92.30}$ \\
 \rbfinf[psd]
 & $96.96 \pm 0.14$ 
 & $\mathbf{85.88 \pm 2.02}$
 & $\mathbf{78.92 \pm 1.91}$
 & $\mathbf{90.70}$ \\
 \end{tabular}
 \caption{Performance of 512-512-512-10 networks for MNIST testing input, and for input corrupted by adversarial attacks computed with perturbation size $\epsilon=0.3$.}
 \label{table-standard-training} 
\end{table*}

\paragraph{Accuracy.}
In Table~\ref{table-standard-training} we summarize the accuracies of networks trained on the MNIST training set, 
as measured both on the (un-perturbed) test set, and upper bounds on the true $\epsilon\!\!=\!\!0.3$-attack accuracy provided by the various adversarial attacks. 
The results are computed as the average of 10 training runs for ReLU and Sigmoid networks, and of 5 runs for \rbfinf\ and \rbfinf[psd].
In each run we used different random seeds for weight initialization; each run consisted of 30 training epochs.
In a result of the form $a \pm e$, $a$ is the percentage accuracy, and $e$ is the standard deviation in the accuracy of the individual runs.

In absence of perturbations, \rbfinf\ networks lose $(1.66 \pm 0.21)\%$ performance compared to ReLU networks (from $(98.62\pm0.07)\%$ to $(96.96\pm0.14)\%$), and perform comparably to sigmoid networks (the difference is below the standard deviation of the results).
When heuristic perturbations are present, the performance of \rbfinf\ networks is superior.
I-FGSM attacks are usually the most effective.

\begin{figure}
\centering
\includegraphics[width=\figsize]{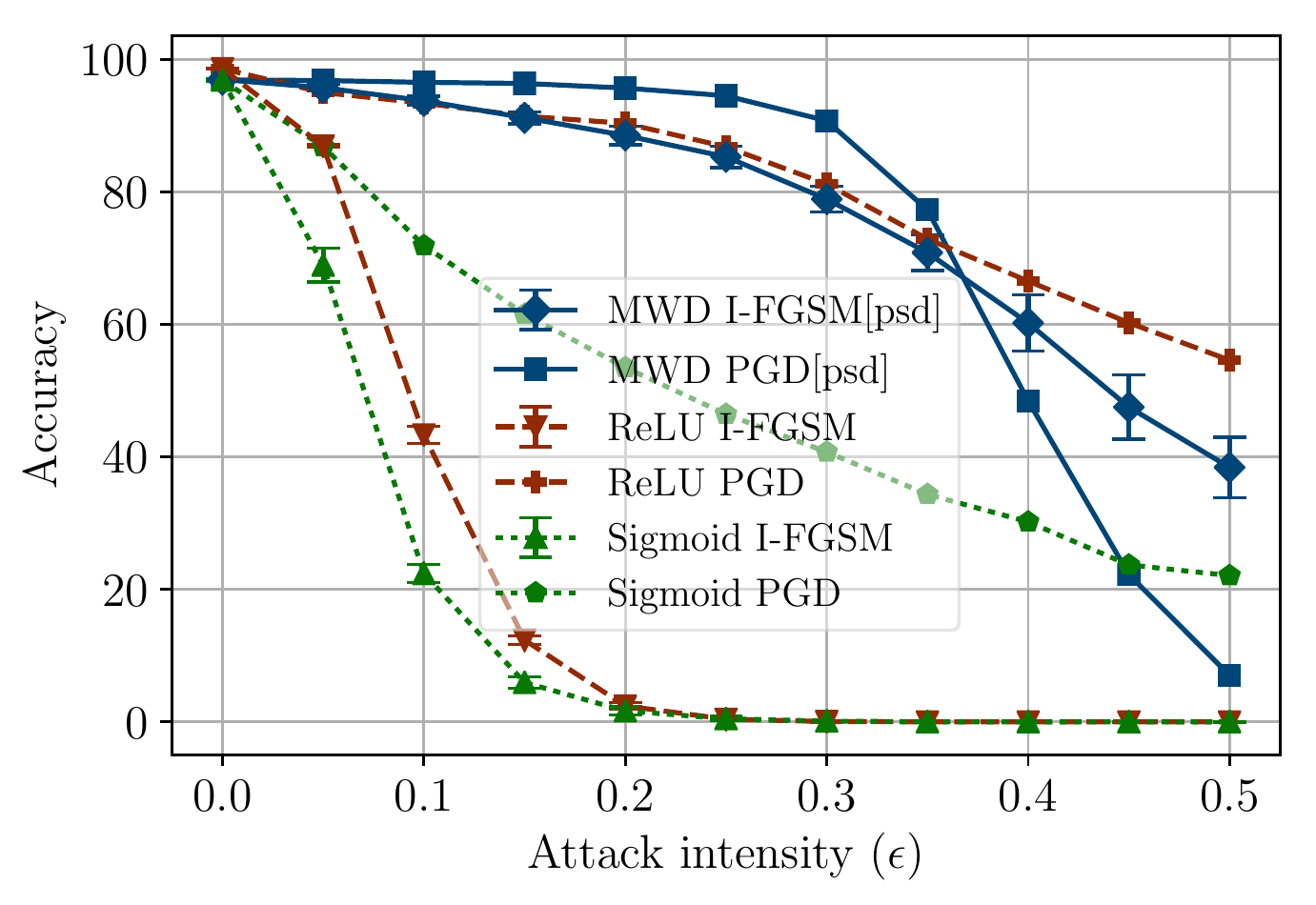}
\caption{Performance of ReLU, Sigmoid, and \rbfinf\ networks in presence of PGD and I-FGSM attacks.}
\label{fig-pgd-ifgsm}
\end{figure}

In Figure~\ref{fig-pgd-ifgsm} we compare the performance of the networks subjected to I-FGSM and PGD attacks for attack amplitudes up to $\epsilon=0.5$. 
The error bars in this and subsequent graphs indicate the standard deviation across the runs, when available.
For \rbfinf\ networks, we plot only attacks based on pseudogradient, as they are more effective. 
We also omitted the results for FGSM, as it is a weaker attack than its iterated version I-FGSM. 
We see that for ReLU and Sigmoid networks, I-FGSM is the strongest attack, giving a rapidly-decaying upper bound on the networks' true $\epsilon$-attack accuracy.
For \rbfinf\ networks, the relative strength of I-FGSM and PGD attacks depends on the intensity $\epsilon$.

\begin{figure}
\centering
\includegraphics[width=\figsize]{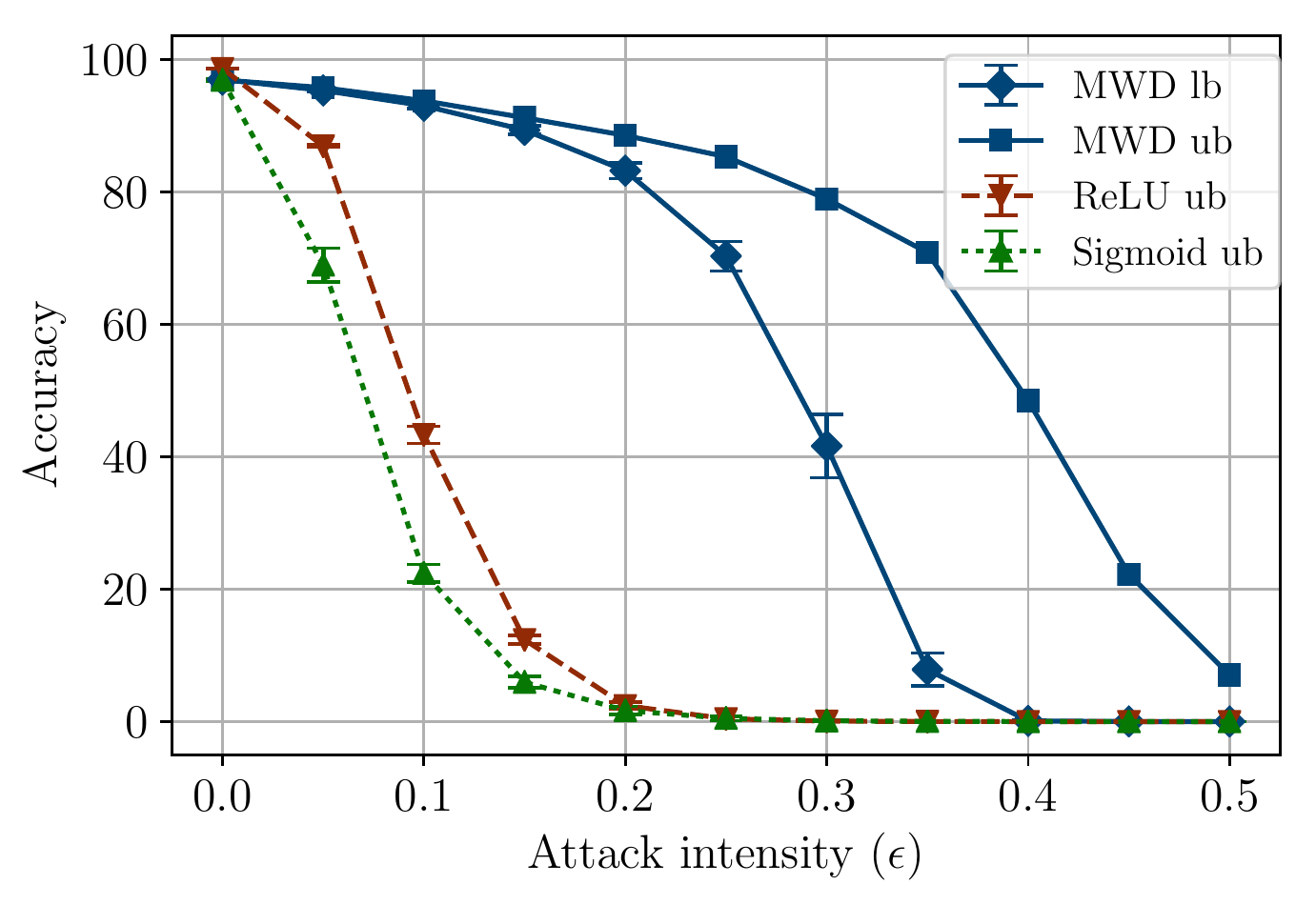}
\caption{Upper and lower accuracy bounds for \rbfinf\ vs. upper accuracy bounds for ReLU and Sigmoid networks.}
\label{fig-bounds}
\end{figure}

\paragraph{Comparing lower bounds for MWD with upper bounds for ReLU, Sigmoid.}
Figure~\ref{fig-bounds} compares the lower bound on the true $\epsilon$-attack accuracy of \rbfinf\ networks with the best upper bounds on the true $\epsilon$-attack accuracy from 
the various attacks on the \rbfinf, Sigmoid, and ReLU networks. 
The lower bound for \rbfinf\ networks is computed with the methods of Section~\ref{sec-bounds}, and holds for all possible adversarial attacks (or perturbations).
The upper bound for \rbfinf\ is derived as the minimum of the curves for PGD[psd] and I-FGSM[psd] in Figure~\ref{fig-pgd-ifgsm}. 
The upper bounds for ReLU and Sigmoid networks are derived from I-FGSM attacks, as they are (in our setting) strictly more powerful than PGD attacks for these networks. 
The upper bounds are not tight: they could be strengthened by performing stronger attacks. 
The lower bound for MWD is also unlikely to be tight since it is based on an approximate analysis (see Section~\ref{sec-bounds}).

Even with these approximations, there is a large accuracy gap between the lower bound guarantee for \rbfinf\ and the upper bounds for ReLU and Sigmoid networks. 
The gap is a visual indication of the stronger resistance to attacks of \rbfinf\ networks. 
For $\epsilon \leq 0.2$, the figure shows that the gap between the upper and lower bounds for MWD is relatively small,
indicating that the lower bounds of Section~\ref{sec-bounds} are fairly accurate in this regime. 
As expected, the bounds go to 0 with increasing $\epsilon$.

We stress that MWD lower bound in Figure~\ref{fig-bounds} gives the {\em provable\/} margin of resistance of \rbfinf\ networks with
respect to adversarial attacks {\em as measured on the testing set.}
If the test set is a representative sample of the inputs that will be seen while the net is in use, then Figure~\ref{fig-bounds} can be interpreted as saying that, 
even with optimally perturbed $\epsilon{=}0.2$-adversarial examples, our \rbfinf\ network still provides an accuracy of over 80\%.  
Of course, this guarantee does not hold if the examples to be predicted on come from a wildly different distribution.
In other words, our accuracy guarantee is conditional to a given input distribution --- as is typical in machine learning.

\paragraph{MWD vs. adversarially-trained ReLU networks.}
Including adversarial examples in the training set is the most common method used to make neural networks more resistant to adversarial attacks \cite{GoodfellowExplainingharnessingadversarial2014,MadryDeepLearningModels2017}.
Therefore, it is interesting to compare the lower bound guarantees for \rbfinf\ networks with the upper bounds on the true $\epsilon$-attack accuracies
 ReLU networks trained via a mix of normal and heuristically generated adversarial examples. 
For brevity, we omit the results for Sigmoid networks, as they were consistently inferior to those for ReLU networks.
Before training on each batch of 100 labeled examples, we add candidate adversarial examples to the batch in one of three different ways.
%
\begin{itemize}
\item {\bf ReLU(FGSM)} and {\bf ReLU(I-FSGM):}
for each $(\vecx, t)$ in the batch, we construct a potential adversarial example $\advvecx$ via (\ref{eq-FGSM}) or (\ref{eq-I-FGSM}), and we feed both $(\vecx, t)$ and $(\advvecx, t)$ to the network for training. 

\item {\bf ReLU(PGD):}
for each $(\vecx, t)$ in the batch, we perform 100 steps of projected gradient descent from a point chosen at random in $B_\epsilon(\vecx) \inters [0,1]^n$; denoting by $\vecx'$ the ending point of the projected gradient descent, we feed both $(\vecx, t)$ and $(\vecx', t)$ to the network for training. 

\end{itemize}
The candidate adversarial examples were generated with $\epsilon=0.3$, which is consistent with \cite{MadryDeepLearningModels2017}.
Due to the high computational cost of adversarial training (and in particular, PGD adversarial training), we performed one run, and we trained the networks for 10 epochs, which was sufficient for their accuracy to plateau.

\begin{figure}
\centering
\includegraphics[width=\figsize]{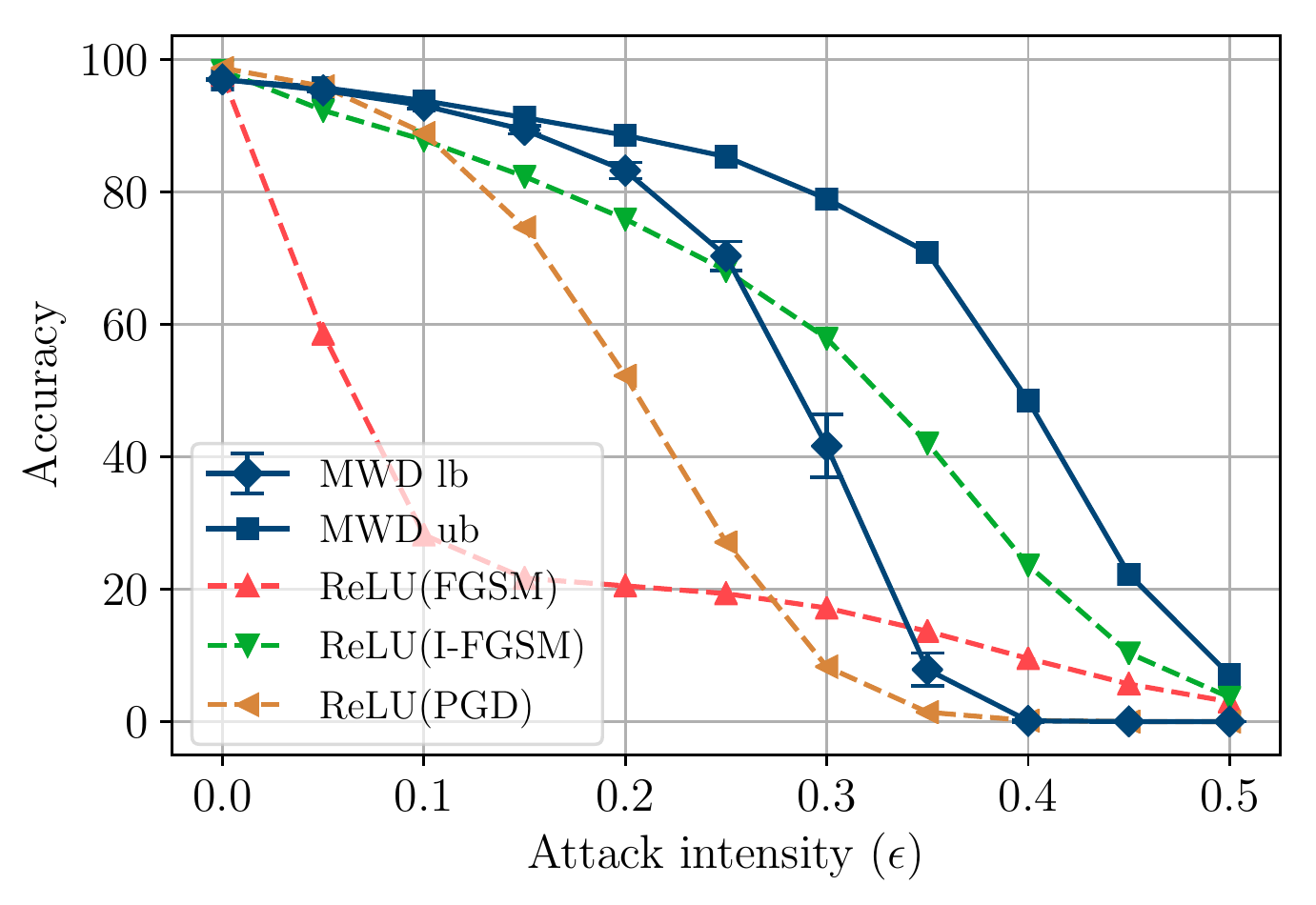}
\caption{Upper and lower accuracy bounds for \rbfinf\ vs. upper accuracy bounds for ReLU networks trained with adversarial examples.}
\label{fig-adversarial-bounds}
\end{figure}

In Figure~\ref{fig-adversarial-bounds} we compare the previously reported accuracy upper and lower bounds for \rbfinf, with the accuracy upper bounds for adversarially-trained ReLU networks. 
The accuracy upper bounds for adversarially-trained ReLU networks are obtained via I-FGSM attacks, which are more effective than PGD or FGSM attacks against such networks.  
We see that the upper bounds for \rbfinf\ networks are above the upper bounds for adversarially-trained ReLU networks for all but the smallest attacks (for $\epsilon \geq 0.05$). 
Furthermore, we see that the lower bounds for \rbfinf\ networks are above the upper bounds for adversarially-trained ReLU networks for a good range of attack sizes ($0.05 \leq \epsilon \leq 0.25$), in spite of both upper and lower bounds being conservative.
Together, this indicates that at least in the training regimes we explored, \rbfinf\ trained without the benefit of adversarial examples offer more resistance to adversarial attacks than ReLU networks trained with the benefit of adversarial examples. 

\paragraph{Training \rbfinf\ networks with pseudogradients vs.\ standard gradients.}

We compared the performance achieved by training \rbfinf\ networks with standard gradients, and with pseudogradients. 
We considered networks with 512, 512, 512, and 10 units, where the first three layers consisted of a random mix of And and Nand units, while the last layer was composed of Nand units (the results do not depend strongly on such unit choices). 
After 30 epochs of training, pseudogradients yielded $(96.79 \pm 0.17)\%$ accuracy, while regular gradients only $(86.35 \pm 0.75)\%$. 
On smaller networks, that should be easier to train, the gap even widened: for networks with 128, 128, and 10 units, pseudogradients yielded $(95.00 \pm 0.29)\%$ accuracy and regular gradients only $(82.40 \pm 3.72)\%$. 
This indicates the need of resorting to pseudogradients for training \rbfinf networks.

	\section{Conclusions and Future Work}

In this paper we advanced the state of the art in producing neural networks resistant to adversarial attacks via two contributions. 
We introduced MWD network units, whose non-linear structure makes them intrinsically resistant to attacks, along with techniques for training networks including MWD units. 
We also provided an efficient technique for computing accuracy guarantees for networks under adversarial attacks, showing that MWD networks provide guarantees that are superior to common network types such as ReLU and sigmoid networks. 

Much work remains to be done, including extending the results to convolutional networks, and exploring the design space of trainable nonlinear structures is clearly an interesting endeavor.

	\bibliographystyle{alpha}
	\bibliography{ml}


\end{document}